*Article*

# Predicting Asphalt Pavement Friction Using Texture-Based Image Indicator

**Bingjie Lu [1, *], Zhengyang Lu [2], Yijiashun Qi [3], Hanzhe Guo [4], Tianyao Sun [5] and Zunduo Zhao [6]**

[1] Independent researcher; bingjie0824@gmail.com
[2] Jiangnan University; luzhengyang@jiangnan.edu.cn
[3] University of Michigan; elijahqi@umich.edu
[4] University of Michigan; hanzheg@umich.edu
[5] Independent researcher; sunstella313@gmail.com
[6] New York University; zz3000@nyu.edu
* Correspondence: bingjie0824@gmail.com

**Abstract:** Pavement skid resistance is of vital importance for road safety. The objective of this study is to propose and validate a texture-based image indicator to predict pavement friction. This index enables pavement friction to be measured easily and inexpensively using digital images. Three different types of asphalt surfaces (dense-graded asphalt mix, open-grade friction course, and chip seal) were evaluated subject to various tire polishing cycles. Images were taken with corresponding friction measured using Dynamic Friction Tester (DFT) in the laboratory. The aggregate protrusion area is proposed as the indicator. Statistical models are established for each asphalt surface type to correlate the proposed indicator with friction coefficients. The results show that the adjusted R-square values of all relationships are above 0.90. Compared to other image-based indicators in the literature, the proposed image indicator more accurately reflects the changes in pavement friction with the number of polishing cycles, proving its cost-effective use for considering pavement friction in mix design stage.

**Keywords:** Pavement friction; Surface texture; Image processing

## 1. Introduction

Skid resistance, the force generated when a tire, prevented from rotating, slides along the pavement surface, is a key indicator for quality management and routine maintenance. Accurately measuring skid resistance ensures that the required friction is achieved before the pavement is put into use, which is crucial for driving safety. Inadequate skid resistance can lead to high risks, such as sliding, rear-end collisions, and longer braking distances, potentially causing traffic accidents [1]. Additionally, efficient measurement techniques are needed to quickly assess whether the pavement in use provides sufficient friction for vehicles [2], facilitating routine management by transportation departments. Furthermore, skid resistance also indicates the degree of wear and aging of the pavement, making accurate prediction a valuable guide for pre-maintenance.

Traditional methods for measuring skid resistance typically involve direct friction measurements. These include field tests, high-speed and laboratory tests, as well as low-speed approaches. High-speed methods encompass four distinct skid resistance measurement techniques, all of which require attaching a trailer or a wheel to a vehicle and measuring the friction force at the interface between a locked, sliding wheel and the wet pavement surface [3], [4]. Laboratory tests often employ the British Pendulum Tester (BPT) and the Dynamic Friction Tester (DFT). These tests measure pavement friction by assessing the loss of kinetic energy in a sliding pendulum or rotating disc when it contacts the roadway surface, converting this loss to a frictional force [5], [6]. Although direct friction measurements provide a clear reflection of the pavement's actual skid resistance, they are limited to spot measurements and are not suitable for large-scale assessments. Additionally, these methods are heavily influenced by the testing vehicle and experimental conditions, necessitating the involvement of skilled technicians. Consequently, researchers have recently proposed new methods for measuring skid resistance.

Recently, novel methods for non-contact measurement of pavement friction have been studied. Pérez-Acebo et al. [7] developed two skid resistance prediction models for an entire road network. These machine learning models consider multiple factors affecting skid resistance and can assist road management departments in predicting future skid resistance. However, this method requires identifying pavement materials and calculating traffic volume, and it does not provide real-time measurements. Roy et al. [8] proposed image-based indicators of microtexture and macrotexture for pavement to predict initial skid resistance. Nonetheless, the predicted skid resistance may not accurately reflect the pavement's condition after years of traffic, as skid resistance can deteriorate over time. Du et al. [9] utilized the signal processing technique of wavelet decomposition to characterize pavement texture but failed to establish a correlation model with friction. Yang et al. [10] employed a deep learning network, FrictionNet-V, based on three-dimensional (3D) laser imaging technology to evaluate skid resistance. However, the extensive and complex information in the raw 3D data can lead to overfitting issues and poor result stability [11, 12], especially when the training sample data is limited, and the network layers are deep. Additionally, the laser equipment required demands high memory for data storage and is expensive [13].

The challenges of friction measurements still exist. Traditional direct friction measurement methods suffer from poor repeatability and low correlation between various equipment indicators. Additionally, these methods can only provide point-based friction predictions, resulting in discontinuous outcomes. Among the novel non-contact methods, texture-based approaches estimate pavement friction by calculating characteristic indicators, but the correlation between these indicators and skid resistance is relatively low. Despite the fact that deep learning has been proven to be a reliable method for estimating pavement friction, its effectiveness is constrained by the quality and size of the training dataset. Inconsistencies and inadequacies in the true value data can confuse the deep learning network during training, leading to inadequate accuracy and low portability to other scenarios [14].

To address these limitations, this study explores one cost-effective and non-intrusive way of using image processing methods to evaluate skid resistance of pavement surface. First, digital images and friction coefficients are collected from asphalt slabs with different surface texture patterns. Then, a series of image processing algorithms are developed. Finally, an interpretable image indicator representing the selected texture feature is proposed to predict friction with good accuracy. Compared to image-based indicators developed in previous studies, the established method can more effectively reflect the changes in pavement friction with varying wear levels, making it potentially applicable to field detection of pavement friction under repeated traffic loading.

## 2. Laboratory Experiment and Data Collection

### *2.1 Friction Data Acquisition*

Laboratory tests were conducted on three asphalt surfaces: Dense Graded Asphalt Concrete (DGAC), Open Graded Friction Course (OGFC), and Chip Seal. Friction coefficients were measured using the Dynamic Friction Tester (DFT). The DFT is a portable instrument designed to measure the friction value of pavement surfaces, allowing for the recording of friction coefficients at various speeds. A three-wheel polisher was used to simulate tire wear on pavement surface. This apparatus consists of a platform with a 68 kN load and three 2.80/2.40-4 tires (35 psi pressure), driven by a motor to rotate at 60 cycles per minute. The polishing process was conducted along a circular wheel path with a diameter of 28.5 cm, which matches the DFT test circle.

The friction data was collected at a constant speed of 40 km/h and recorded after the vehicle tire had rotated 0k, 50k, 90k, 150k, 300k, 390k, 450k, and 500k times on the samples. On the same sample, friction can vary slightly at different positions due to the effects of water, load, and other factors [15]. Therefore, all pavement slabs were kept dry prior to conducting the experiment. Additionally, external conditions such as temperature and humidity were kept consistent during data collection to minimize the impact of environmental factors. The evolution of friction coefficients with polishing cycles is shown in Figure 1, respectively, for three different asphalt surfaces.

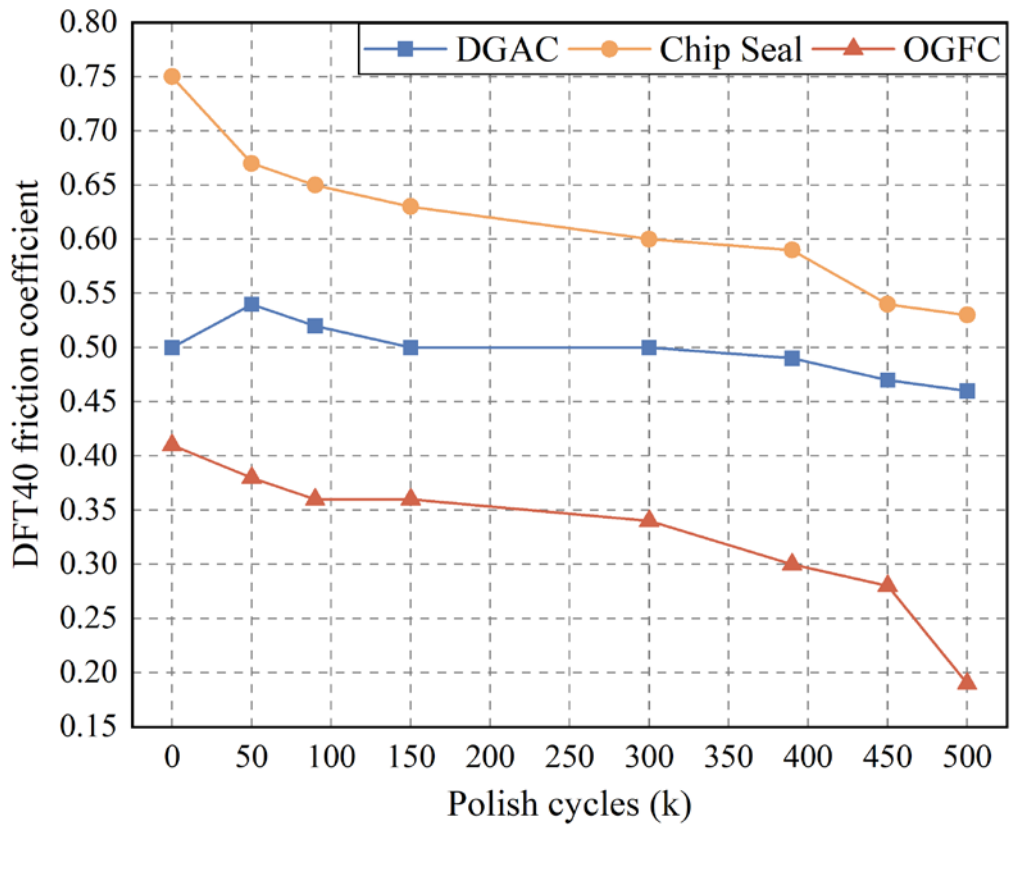

**Figure 1.** DFT40 vs. Polish cycles for different mixtures.

### *2.2 Digital Images Acquisition*

Images of pavement surface were captured when the polisher had abraded the surface to the set number of rotations. The images were taken using a digital camera positioned to face the asphalt slab directly, with a resolution of 350 dpi for both horizontal and vertical dimensions and a pixel size of 6000×4000. Each image contained 24,000,000 pixels, with each pixel consisting of three values (R, G, B). Considering that the quality of pavement image texture is mainly determined by factors such as lighting conditions, camera position, and other variables [16], at least 10 images with 8 different light source angles (Figure 2) were collected for each level of wear to evaluate the practicality of the proposed image processing methods. The images were taken with the corresponding friction values and polishing cycles, some examples are shown in Figure 3.

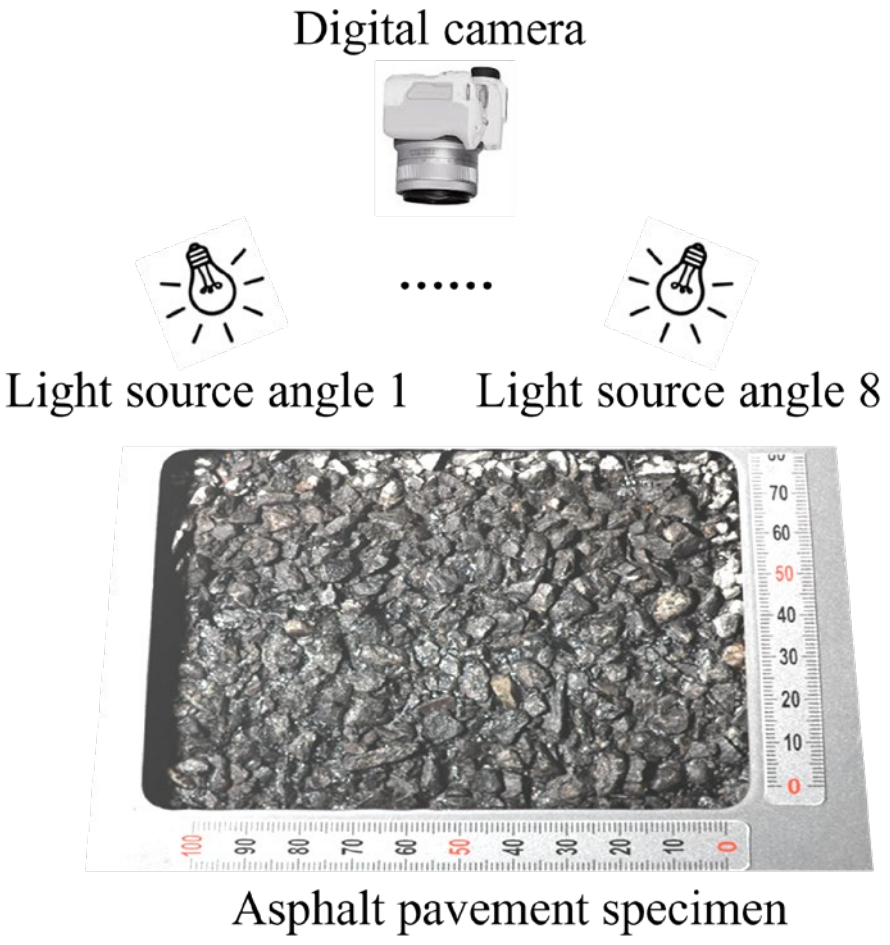

**Figure 2.** Digital images taken with different lighting angles.

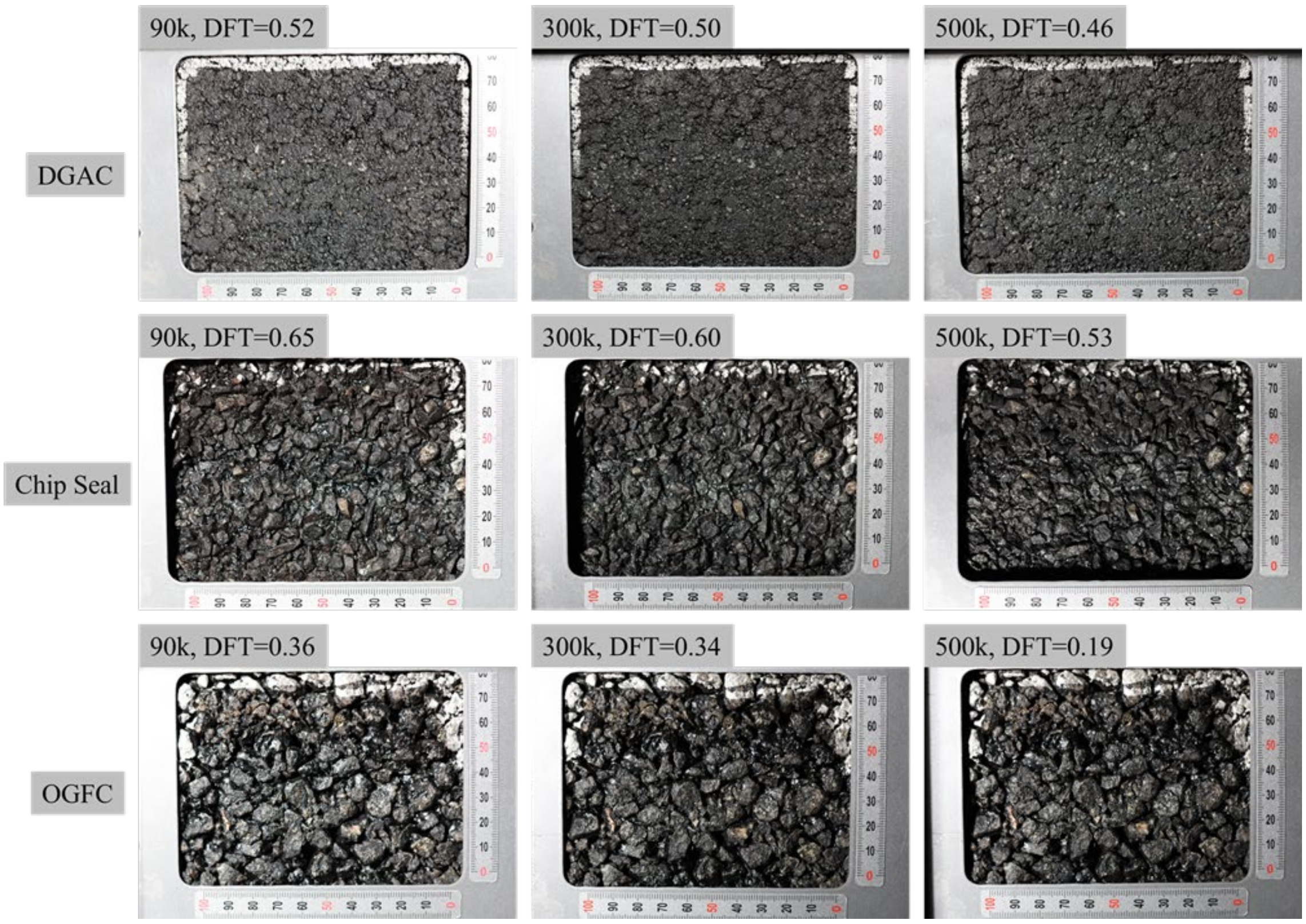

**Figure 3.** Raw images of pavement surface with corresponding DFT40 friction coefficients and polishing cycles.

## 3. Analysis Methodology

### *3.1 Image Preprocessing*

A series of image processing algorithms were performed to obtain the binary image and extract the image-based texture feature for friction prediction, as shown in Figure 4.

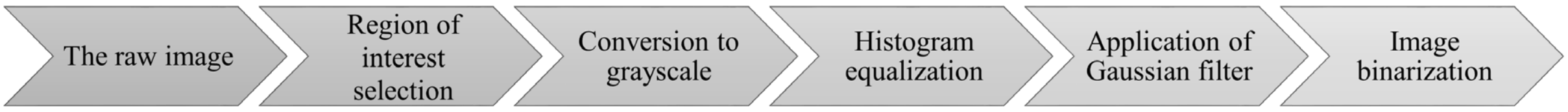

**Figure 4.** Flowchart for pavement texture feature extraction.

Frist, a region of interest (ROI) sized 100 mm × 75 mm is selected and cropped from the color image (as shown within the red box in Figure 5(a)). This step removes the reflective silver calibration board and marks white areas of the sample, as these areas do not represent the true brightness values of the pavement to be analyzed. In practice, users of this method can adjust the camera's height above the ground according to the area of the pavement they wish to measure, thus eliminating the need for cropping. The equipment used in this study does not have a fixed position, resulting in slight variations in the distance between the lens and the sample each time an image is captured. To mitigate the impact of inconsistent shooting height, each cropped image is adjusted to the same pixel dimensions of 3400 × 2550, as shown in Figure 5(b). The image resolution is 0.0294 × 0.0294 mm, which meets the required precision for pavement skid resistance analysis [17]. Next, the adjusted RGB image is converted to an 8-bit grayscale image using the weighted average method [18]. The calculation method is expressed in Equation 1:

$$Gray = 0.299 \times R + 0.587 \times G + 0.114 \times B \quad (1)$$

Where, Gray is the resulting grayscale value, and R, G, and B are the red, green, and blue channel values of the original image, respectively.

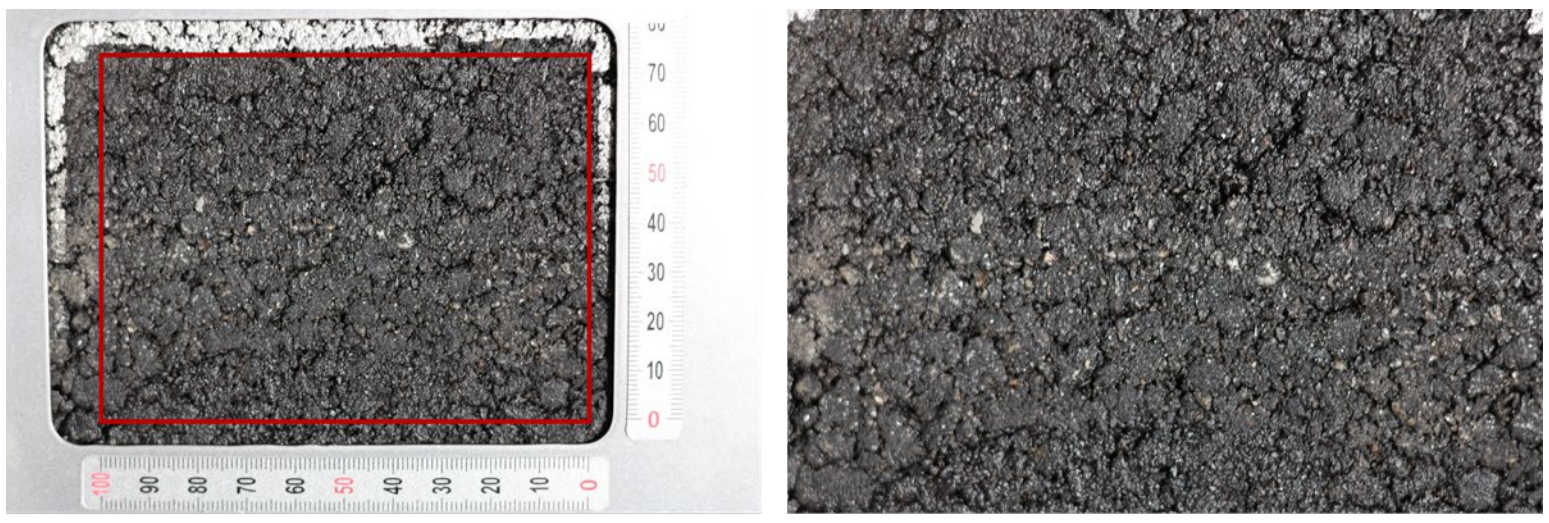

(a) (b)

**Figure 5.** ROI selection: (a) the raw image, (b) the cropped and resized image.

In practice, shooting devices occasionally generate unqualified images due to the impact of external light, surface reflectance properties, and other factors [19, 20]. As shown in Figure 6, three representative images are selected from the image sets of each mixture type to demonstrate these effects. The original photos show the imaging results under different lighting conditions and material types. For instance, the third original photo in the DGAC category illustrates that when the light incidence angle is not perpendicular to the pavement, part of the pavement appears clear while another part is shrouded in shadows. Similarly, the second original photo in the OGFC category demonstrates that when the pavement color is darker and the light source is tilted, it only illuminates a small portion of the pavement. However, in the first original photo of the OGFC category, even if the lighting angle is orthogonal, the reflective properties of the pavement can cause the aggregates to exhibit uneven brightness. Since the subsequent threshold selection step performs binarization based on pixel brightness values, the brightness distribution of such unqualified images needs to be adjusted to ensure consistency for determining a uniform threshold for each type of mixture.

Typically, histogram equalization techniques are used to redistribute the grayscale values of an image to address such issues. However, traditional methods may result in the loss of local details, especially in images with varying brightness [21]. Adaptive Histogram Equalization (AHE) mitigates this by dividing the image into multiple small blocks (referred to as 'tiles' or 'grids') and applying histogram equalization to each tile individually [22]. The boundaries between tiles are then smoothed using bilinear interpolation. Building on AHE, Contrast Limited Adaptive Histogram Equalization (CLAHE) introduces the concept of contrast limiting [23]. This involves clipping the histogram of each tile, with the clip limit being a predefined parameter that determines the maximum number of pixels for any single grayscale level [24]. This prevents excessive contrast enhancement and avoids amplifying noise. As shown in Figure 6, CLAHE is applied to the grayscale image by transforming the values to achieve a brightness distribution where all values are equally probable. The original images under different lighting conditions show varying grey distributions. After applying CLAHE, the grey levels of the same type of pavements are distributed equally. This eliminates discrepancies in the binarization results caused by varying lighting conditions, leading to more consistent texture feature values. Additionally, this algorithm enhances the local contrast of the image, making the details more pronounced.

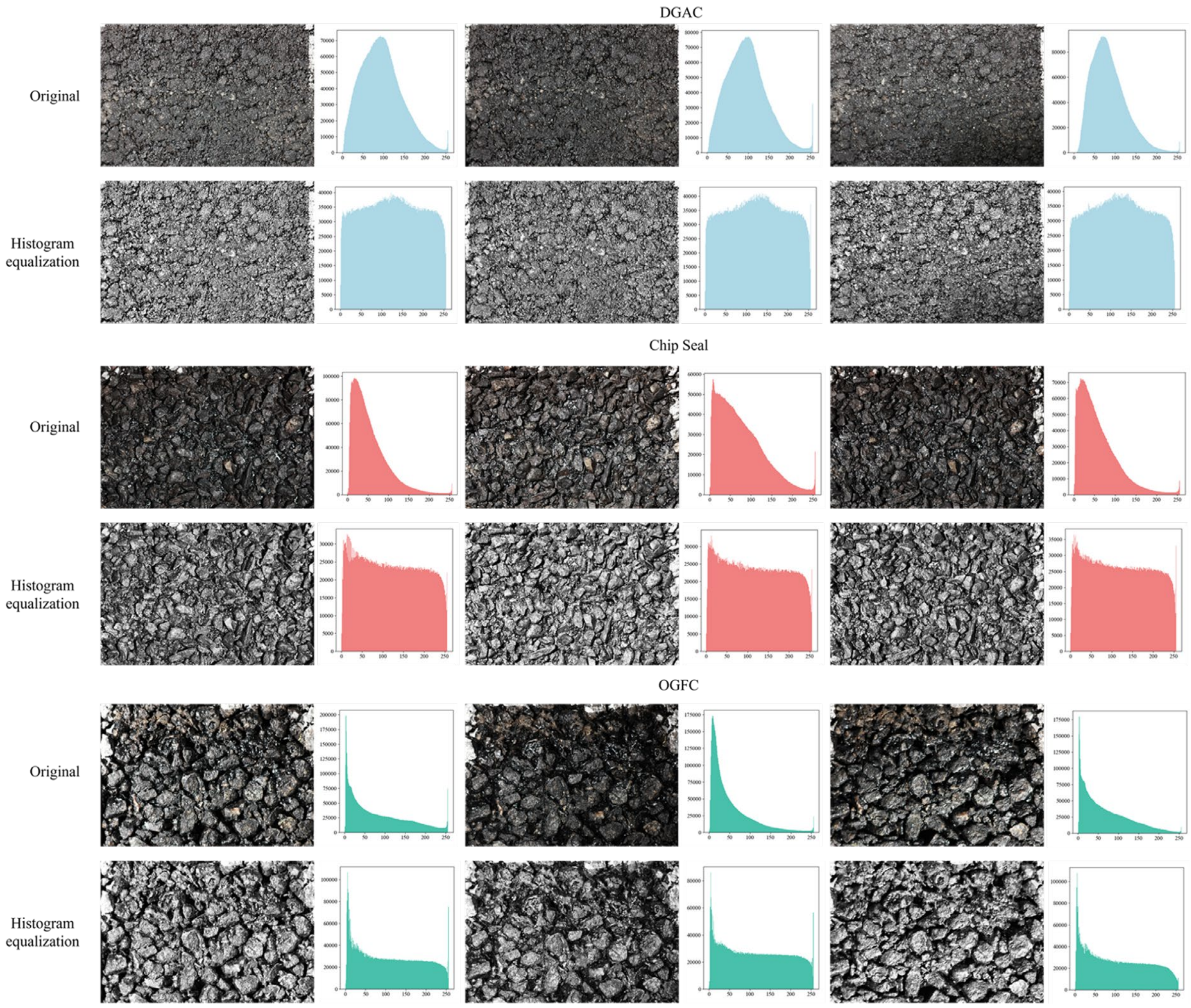

**Figure 6.** The comparison before and after CLAHE.

For subsequent image segmentation tasks, the histogram-equalized image needs to be denoised. Due to its smooth and artifact-free characteristics, the Gaussian filter is widely used [25, 26]. A two-dimensional (2D) Gaussian filter is chosen to smooth the surface. This filter smooths the image through a convolution operation, with the convolution kernel generated by the 2D Gaussian function G(x,y). The calculation method is defined in Equation 2:

$$G(x,y)=\frac{1}{2\pi\sigma^2}e^{-\frac{x^2+y^2}{2\sigma^2}} \tag{2}$$

Where, (x,y) are coordinates relative to the center of the Gaussian kernel, and $\sigma$ is the standard deviation that determines the width of the Gaussian distribution. When the filter smooths the image, pixels in different positions in the neighborhood are given different weights. Pixels farther from the center of the Gaussian kernel (filter window) have smaller weights [27]. Thus, more of the overall gray distribution characteristics of the image are retained compared to median and arithmetic mean filters [28].

*3.2 Texture Feature Extraction*

The contact area between the tire and the road surface significantly impacts the skid resistance of the pavement [29]. To extract the area metric, we can assume the existence of a frictional contact surface. Textures above this surface will come into contact with the rubber and generate friction, while textures below this surface will have little to no contact with the rubber. If 3D laser data of the pavement is available, the position of this contact surface relative to the pavement can be calculated, and then the actual contact area can be predicted using methods such as machine learning [30]. However, these methods require collecting a large amount of data to train the machine learning model. Additionally, regression models need to be established to achieve the final goal of predicting friction [31], making the entire prediction process computationally expensive and time-consuming.

Due to the significant impact of the aggregate's morphological characteristics on the surface properties of asphalt pavement [32], and the fact that aggregates with sharp, angular shapes offer better skid resistance [33], it is concluded that the more protruding aggregates the pavement has, the stronger the friction the tire will experience. Therefore, even though 2D images cannot provide depth information like 3D point cloud data and we cannot calculate the absolute elevation of the contact surface to extract realistic contact area, we can still quantify texture features based on the difference in grayscale information between the protruding parts of the aggregates and both the non-protruding parts of the aggregates and the asphalt. Since the grayscale distribution of images of the same mixture type is adjusted for consistency in the preprocessing step, and the grayscale contrast between the protruding aggregates and other parts of the pavement is enhanced, an appropriate binarization algorithm can determine a unified threshold for each pavement type to effectively segment the grayscale image, as shown in Figure 7, retaining only the aggregates protruding from the surface (marked in black) and removing other regions by treating them as background (marked in white). For the sake of convenience in the text, this textural feature will be referred to as "Area".

In the field of image processing, researchers have proposed various binarization methods, which are mainly divided into global thresholding methods and local (adaptive) thresholding methods [34]. This study opts for the global thresholding method, primarily because the defined protruding part of the aggregates is relative to the entire pavement, rather than relative to its surrounding small area. Consequently, a uniform threshold should be employed to the entire image instead of determining the threshold at each pixel location based on the pixel value distribution of its neighborhood block. Representative global thresholding methods like Otsu's Method [35] and IsoData [36] have demonstrated robust performance in image segmentation. Considering the characteristics of the grayscale distribution of the analyzed images, this paper selects the IsoData method, which adapts better to complex gray-level distributions and is less sensitive to noise. Otsu's method is not chosen because it is more suitable for bimodal gray-level histograms and performs poorly with unimodal or multimodal distributions [37].

Specifically, the IsoData method selects an initial threshold T0 (the mid-value of the grayscale range). The image is then split into a low grayscale part (grayscale value ≤ T) and a high grayscale part (grayscale value > T). The mean grayscale value $\mu_L$ for the low grayscale part (Equation 3) and the mean grayscale value $\mu_H$ for the high grayscale part (Equation 4) are calculated. The threshold is updated according to Equation 5. The difference between the new threshold $T_{new}$ and the old threshold $T$ is compared. If the difference is less than the preset tolerance ε, the threshold is considered to have converged, and the final threshold $T_{final}$ is used to convert the grayscale image into a binary image (Equation 6). If not, the new threshold $T_{new}$ is set as the current threshold $T$, and the iterative calculation continues.

$$\mu_L = \frac{1}{N_L}\sum_{i=1}^{N_L} I_i \tag{3}$$

$$\mu_H = \frac{1}{N_H}\sum_{i=N_L+1}^{N_L+N_H} I_i \tag{4}$$

$$T_{new} = \frac{\mu_L + \mu_H}{2} \tag{5}$$

$$I(x,y) = \begin{cases} 1, \text{ if } I_{gray}(x,y) > T_{final} \\ 0, \text{ if } I_{gray}(x,y) \le T_{final} \end{cases} \tag{6}$$

Where, $N_L$ and $N_H$ are the total number of pixels in the low and high grayscale parts, respectively. $I_i$ represents the grayscale value, and $I$(x, y) is the pixel value of the binary image at the coordinates (x, y).

As previously mentioned, by adjusting the grayscale distribution of each type of mixture to be consistent, the thresholds derived from this algorithm for various pavement images are relatively similar. The thresholds set for the images of DGAC, Chip Seal, and OGFC surfaces are 127, 124, and 115, respectively. Users should note that due to differences between field operation conditions and the experimental setup in this study, such as lighting conditions during photography and the color of the mixtures, the thresholds used in this paper should be further calibrated for practical use. The primary purpose of this study is to provide an imaging-based methodology for quickly estimating pavement friction. More extensive data can be utilized to further train and refine the models established in this study.

Some of the segmentation results are shown in Figure 7. For the same type of mixture, the area of the protruding aggregates decreases after more polishing cycles, corresponding to a reduction in the actual friction [38]. Given that the number of pixels along the width of the image is 2550 and the corresponding actual length is 75 mm, it is straightforward to calculate the actual area size of the black parts in the binarized image. Table 1 presents some of the calculated results of the Area indicators along with their corresponding true friction values and the number of polishing cycles. It is evident that the Area on the same type of pavement shows a regular linear change with friction. However, the pattern for any one type of pavement cannot be directly applied to others. Therefore, it is necessary to develop separate models for different pavement types. Detailed experimental results can be found in the RESULTS AND DISCUSSION section. More variables are not introduced because overfitting is a common issue in multiple linear regression models when the sample size is small relative to the number of predictors [39]. Additionally, capturing the nonlinear relationships between the predictors and DFT is challenging with limited data [40].

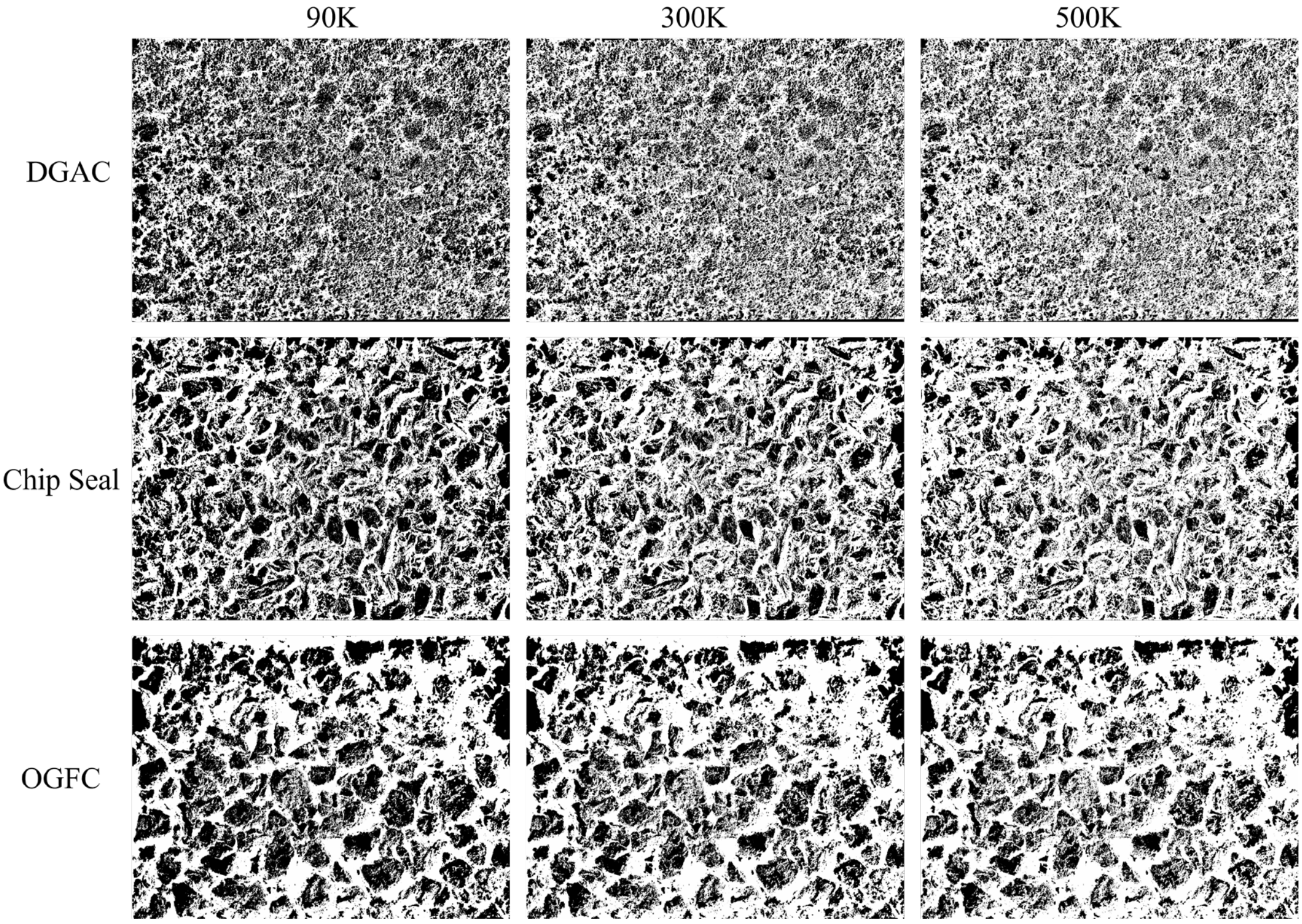

**Figure 7.** Binarization results of different mixtures after varying numbers of polishing cycles.

**Table 1.** Examples of texture feature calculation results.

| Pavement Type | Area (mm²) | Friction | Polishing Cycles (k) |
|---|---|---|---|
| DGAC | 2795.53 | 0.54 | 50 |
| | 2611.88 | 0.52 | 90 |
| | 2427.15 | 0.50 | 150 |
| | | …… | |
| Chip Seal | 2547.03 | 0.67 | 50 |
| | 2122.78 | 0.63 | 90 |
| | 1856.24 | 0.60 | 150 |
| | | …… | |
| OGFC | 2518.15 | 0.38 | 50 |
| | 2634.07 | 0.36 | 90 |
| | 2456.65 | 0.36 | 150 |
| | | …… | |

*3.3 Other Image-Based Indicators from Literature*

Various image-based indicators have been proposed to directly or indirectly predict the pavement friction. This study selects three indicators that have demonstrated relatively good predictive performance in previous research experiments. These indicators are used to establish relationships with DFT friction to evaluate the effectiveness and superiority of the proposed texture feature.

Valikhani et al. [41] proposed an image-based metric called Aggregate Ratio (AR), defined by Equation 7:

$$AR = A_{aggregate}/A_{surface} \tag{7}$$

Where, $A_{surface}$ represents the area of the surface, which is equal to the entire image size, and $A_{aggregate}$ represents the complete particle area of aggregates. The selection of image thresholds in their study is based on the grayscale value differences between the aggregate surface and the black asphalt. In contrast, our study determines thresholds based on the differences in grayscale values between the protruding parts of the aggregates and both the non-protruding parts of the aggregates and the asphalt.

Roy et al. [8] utilized the wavelet transform method to extract image texture features. By decomposing the image into sub-images of different resolutions and detail images containing high-frequency details, detail coefficients are extracted from the high-frequency detail images. The energy features are obtained by calculating the arithmetic mean of the squared values of the detail coefficients at each decomposition level. Based on these energy features, a macro-texture indicator called the Surface Macrotexture Index (SMI) is proposed to determine the skid resistance of newly constructed road surfaces.

Wan et al. [42], based on the brightness distinction between convex and concave parts, used the maximum entropy method to segment the concave areas of the pavement. They then employed the Fractal Dimension (FD) to characterize the concave distribution characteristics (CDC) of pavements, establishing a strong correlation with texture depth (TD) and mean texture depth (MTD) obtained through the sand patch method. The FD is calculated using the box-counting method, as defined in Equation 8.

$$FD = -\lim_{\varepsilon \to 0} \frac{\log(N(\varepsilon))}{\log(\varepsilon)} \tag{8}$$

Where, ε represents the changeable box size and $N\varepsilon(P)$ represents the minimum number of an n-dimensional box. The calculation processes of all the above indicators are reproduced according to the previous literature [41], [8], [42], where the calculation details can be found.

## 4. Results and Discussion

### *4.1 Evaluation of Proposed Image Indicator*

The study is conducted on a laptop with an AMD Ryzen 9 5900 HX central processing unit (CPU). Data from DGAC, Chip Seal, and OGFC surfaces in the dataset are used to build the relationship models. As shown in Figure 8, the Area determined from image analysis is compared to DFT data to develop separate models for different pavement types. Most of the data points are found close to the line of equality, indicating that the proposed indicator is valid. As the number of times the tires polish the pavement increases, friction is expected to reduce as the macrotexture diminishes and the area of protruding textures decreases. However, as documented in [43], the friction of the pavement can slightly increase in the early stages of grinding. At this stage, if one touches the pavement with a finger, the texture appears rougher and potentially more abrasive. Except for DGAC pavement, our data does not reflect this friction change trend in the early stages, primarily because the intervals between the polishing cycles when measuring the friction were not small enough. Research indicates that the macrotexture of aggregate surfaces significantly impacts the skid resistance of pavement, and this macrotexture is primarily controlled by aggregate gradation parameters [44]. Generally, mixtures with a larger nominal maximum aggregate size (NMAS) tend to provide better skid resistance [45]. As shown in Figure 8, Chip Seal exhibits higher surface friction compared to DGAC. However, this evaluation criterion is not absolute, as the friction of OGFC is lower than that of DGAC.

Table 2 summarizes the indicators of the linear regression equations, and the calculation formulas for the three selected indicators are shown in Equation 9–11:

$$r = \frac{\sum (y_i - \bar{y})(\hat{y}_i - \bar{\hat{y}})}{\sqrt{\sum (y_i - \bar{y})^2 (\hat{y}_i - \bar{\hat{y}})^2}} \tag{9}$$

$$R^2 = 1 - \frac{\sum (y_i - \hat{y}_i)^2}{\sum (y_i - \bar{y})^2} \tag{10}$$

$$R_{adj}^2 = 1 - \left(\frac{1 - R^2}{n - p - 1} \times (n - 1)\right) \tag{11}$$

Where, r represents the Pearson Correlation Coefficient, which measures the linear relationship between two variables. $R^2$ is the Coefficient of Determination (COD) [46], representing the proportion of variance in the dependent variable that is predictable from the independent variable. $R^2_{adj}$ is the Adjusted Coefficient of Determination, which adjusts the $R^2$ value based on the number of predictors in the model, providing a more accurate measure of the model's explanatory power. yi are the observed values, and $\hat{y}_i$ are the predicted values. $\bar{y}$ is the mean of the observed values, and $\bar{\hat{y}}$ is the mean of the predicted values. n is the number of observations, and p is the number of predictors [47, 48].

The adjusted $R^2$ values for all three models are greater than 0.90, indicating significant relationships [49]. The different slopes of the three regression lines suggest that the rate of friction reduction varies among different pavement types under the same wear conditions. For example, the friction reduction of OGFC pavements is significantly higher than that of DGAC pavements. This information can guide the selection of materials for pavements with different functions. For instance, DGAC material is a better choice for constructing long-lasting pavements.

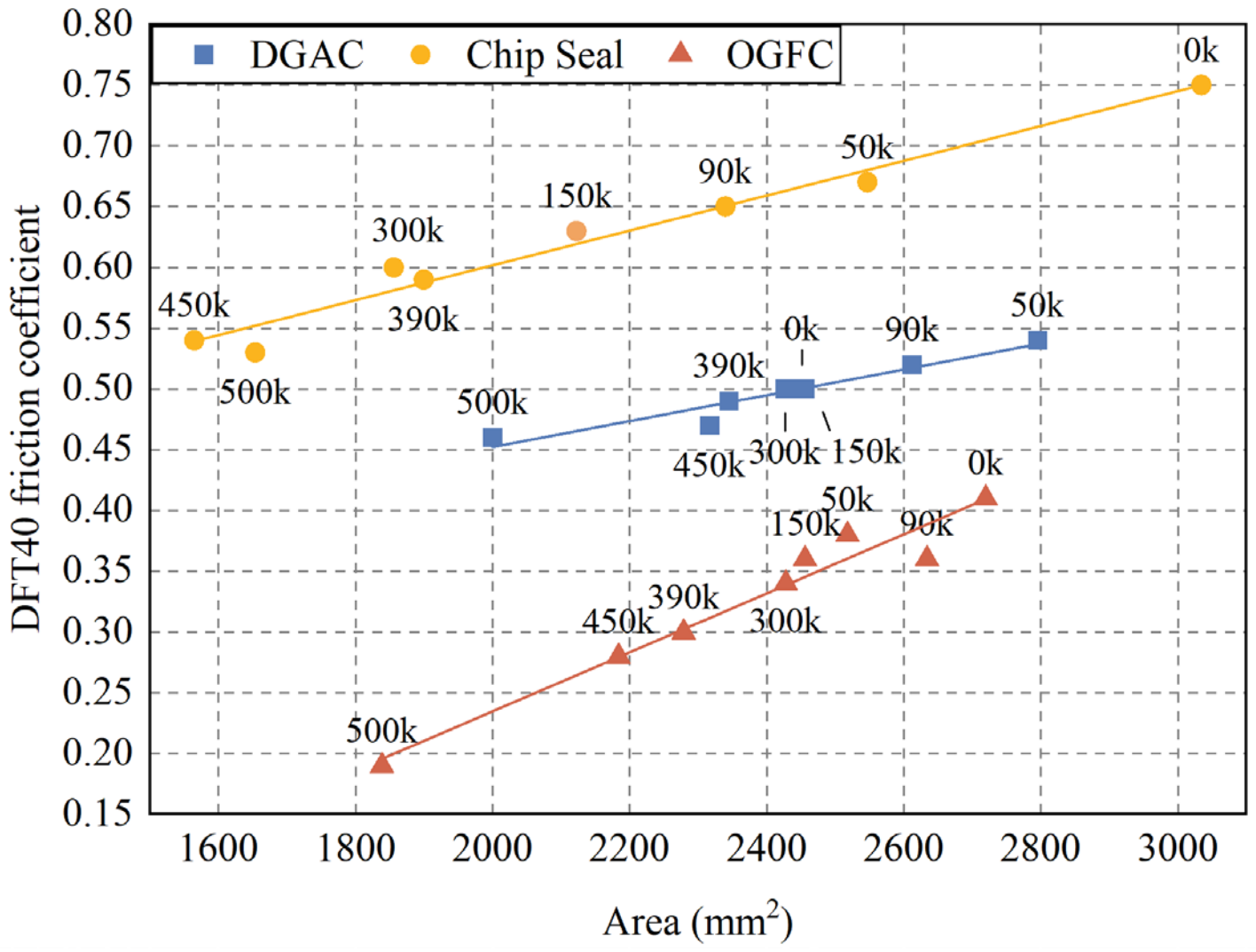

**Figure 8.** Relationships between the proposed image indicator and DFT40 friction coefficients for three asphalt surfaces.

**Table 2.** The established friction prediction models for various types of pavements.

| Pavement type | Pearson's correlation | $R^2$ | $R^2_{adj}$ | Regression models |
|---|---|---|---|---|
| DGAC | 0.9620 | 0.9255 | 0.9130 | DFT= 0.2396+1.0632E-4Area |
| Chip Seal | 0.9851 | 0.9704 | 0.9655 | DFT= 0.3151+1.4331E-4Area |
| OGFC | 0.9782 | 0.9569 | 0.9498 | DFT=-0.2504+2.4260E-4Area |

Regarding the sources of error in the proposed prediction method, the following three possibilities are identified: First, the limited number of samples leads to limited accuracy in prediction models. Second, obtaining the actual friction values of the pavement under high-speed conditions is challenging, which also results in some inevitable prediction errors. Third, the threshold selection method for image binarization is not perfect, resulting in some texture features being ignored or overly enhanced.

*4.2 Comparison of Accuracy for Different Image-Based Indicators*

The same image dataset was used to calculate three image-based indicators in the literature (AR, SMI, and FD) and evaluate their accuracy for friction prediction. Figure 9 shows the correlation between the AR calculation results and DFT friction obtained using linear regression. Compared to the proposed indicator, the $R^2$ values of the three fitted lines are quite low, indicating they do not effectively capture the trend of the data points. This is because, using the threshold-based method from the literature, the area of exposed aggregates on the same pavement does not significantly change with different polishing cycles. The minor variations in AR values for the same mix under different polish cycles are due to the contrast and brightness of the images not being perfectly adjusted to the same levels. In other words, the proposed indicator can better simulate the interaction between aggregates and tires as they undergo repeated polishing, providing a more accurate prediction of friction. The AR indicator might be more suitable for evaluating the initial friction of different types of pavements, offering a simple and effective means for approximate friction assessment. However, due to the limited number of mix types included in this study, we did not establish a predictive model to verify the potential applicability of this indicator.

Figure 10 shows the results of fitting SMI and DFT using linear equations for different mixture types. The adjusted $R^2$ values for the DGAC and Chip Seal models are both higher than 0.90, indicating a significant relationship between SMI and DFT in this case, and SMI performs similarly to the proposed indicator. Due to the wavelet transform method's superior noise resistance compared to the threshold segmentation method, it allows for more accurate extraction of macrotexture features. Therefore, extracting the SMI to represent pavement friction is even more effective, as demonstrated in the case of DGAC. The adjusted $R^2$ value between SMI and DFT is 0.9481, while the adjusted $R^2$ value between Area and DFT is 0.9130. However, the adjusted $R^2$ value for the OGFC model is only 0.8061, indicating a weaker correlation between SMI and DFT, which is worse than the proposed indicator. Considering that friction is influenced by both macrotexture and microtexture, this indicates that the contribution of microtexture to OGFC friction cannot be ignored. The macro-texture indicator (SMI) fails to fully capture this influence because wavelet transform involves complex multi-scale analysis, which may lead to microtexture information loss or the introduction of errors during feature extraction. In contrast, threshold segmentation, which directly processes the grayscale values of the images, is relatively simple and can better preserve the original information.

Figure 11 illustrates the relationship between FD and DFT for various mixtures. According to the adjusted $R^2$ values of the models, using FD to predict the friction of Chip Seal performs similarly to the proposed indicator but performs worse for DGAC and OGFC. This is because FD can capture the complexity and irregularity of surfaces at different scales, making it excellent at representing the texture of Chip Seal pavements, which have rough surfaces with significant multi-scale irregularities. For DGAC pavements, which have relatively regular and smooth textures, the FD does not provide additional useful information. In this case, a simple threshold selection method can more effectively capture the key surface features. For OGFC, a porous hydrophobic material, image-based indicators fail to capture the internal pore structure of such pavements. Compared to concave distribution characteristics (CDC), the protruding aggregates have a more direct and significant impact on friction performance. Additionally, the FD method requires the selection and tuning of multiple parameters, such as the choice of wavelet basis and the number of decomposition levels. Incorrect parameter selection can lead to poor prediction results. In contrast, the proposed method is relatively easy to implement, requiring only the selection of an appropriate threshold based on the type of mixture.

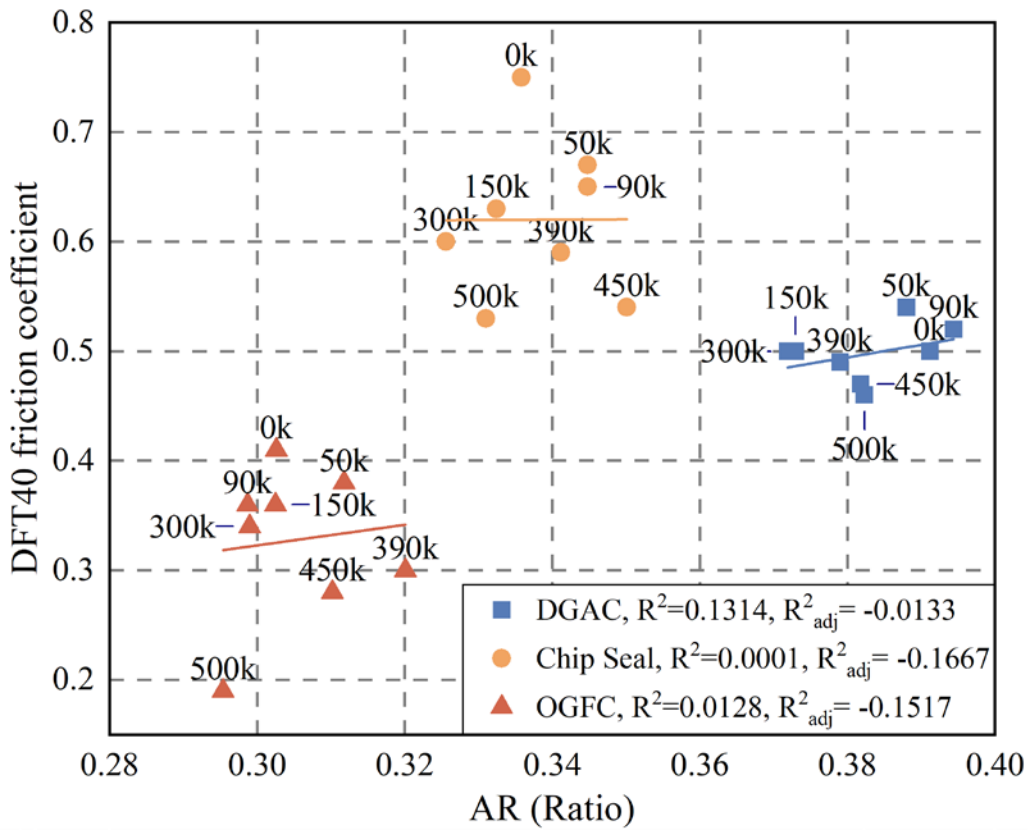

**Figure 9.** Relationship between AR and DFT40 friction coefficients for three asphalt surfaces.

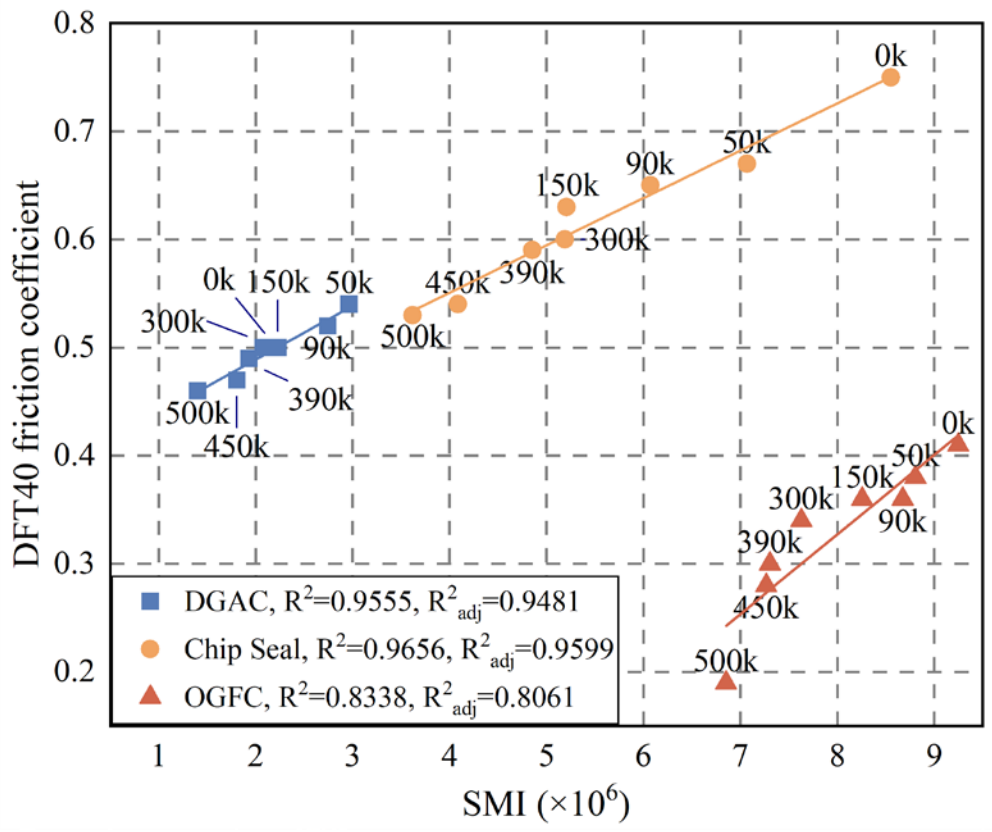

**Figure 10.** Relationship between SMI and DFT40 friction coefficients for three asphalt surfaces.

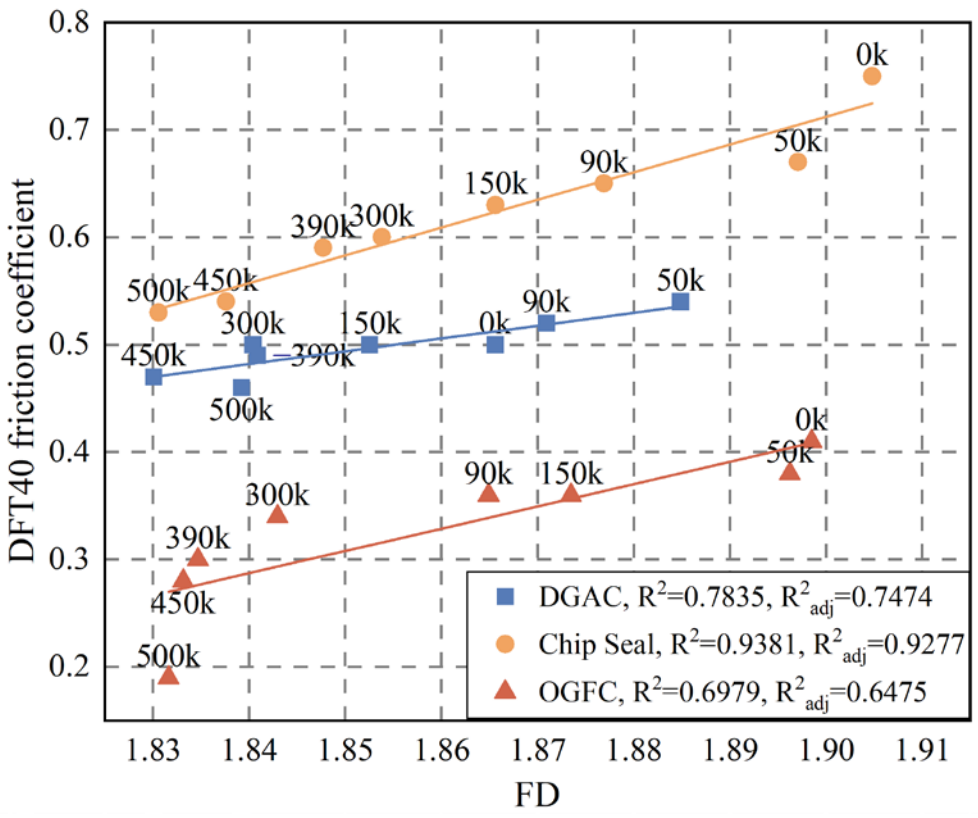

**Figure 11.** Relationship between FD and DFT40 friction coefficients for three asphalt surfaces.

## 5. Conclusions

This study proposed a texture-based image indicator for characterizing surface texture feature and predicting pavement friction. The proposed image preprocessing techniques effectively mitigate the effects of varying lighting conditions and camera heights during photography. Modeling results indicate that the aggregate protrusion area, extracted using a threshold selection algorithm, establishes a strong correlation with DFT-measured friction coefficients. The adjusted R2 values of friction prediction models exceed 0.90 for each of three different types of pavement surfaces. Comparative analysis demonstrates that the proposed image indicator outperforms other image-based indices, particularly in accurately reflecting changes in pavement friction with polishing cycles.

It is proved that the proposed image indicator and models can be used to evaluate pavement friction during mix design phase. However, it is important to note that this study developed prediction models for each type of asphalt mixture or surface treatment separately, and the established method may not be applicable to pavement surface types not included in the dataset for model development. Future work is needed to evaluate the feasibility of image indicator for network application using images taken from the moving vehicle and friction measurements from the skid tester in the field.

**Author Contributions:** Conceptualization, Bingjie Lu; methodology, Bingjie Lu; validation, Bingjie Lu and Zhengyang Lu; formal analysis, Bingjie Lu; writing—original draft preparation, Bingjie Lu; writing—review and editing, Zhengyang Lu, Hanzhe Guo; visualization, Yijiashun Qi, Tianyao Sun, Zunduo Zhao. All authors have read and agreed to the published version of the manuscript.

**Funding:** This research received no external funding.

**Data Availability Statement:** The data presented in this study are available on request from the corresponding author. The data are not publicly available due to privacy.

**Conflicts of Interest:** The authors declare no conflicts of interest.